\documentclass[10pt,twocolumn,letterpaper]{article}

\usepackage{wacv}
\usepackage{times}
\usepackage{epsfig}
\usepackage{graphicx}
\usepackage{amsmath}
\usepackage{amssymb}
\usepackage{multirow}
\usepackage{booktabs}

\usepackage{algorithm}
\usepackage{algorithmic}
\renewcommand{\algorithmicrequire}{\textbf{Input:}}
\renewcommand{\algorithmicensure}{\textbf{Output:}}

\def\wacvPaperID{651} 

\wacvfinalcopy 

\ifwacvfinal
\def\assignedStartPage{9876} 
\fi

\ifwacvfinal
\usepackage[breaklinks=true,bookmarks=false]{hyperref}
\else
\usepackage[pagebackref=true,breaklinks=true,colorlinks,bookmarks=false]{hyperref}
\fi

\ifwacvfinal
\else
\fi

\begin{document}

\title{Uncertainty-aware Mean Teacher for Source-free \\Unsupervised Domain Adaptive 3D Object Detection}




\author{ Deepti Hegde, Vishwanath Sindagi,Velat Kilic, A. Brinton Cooper, Mark A. Foster and Vishal M. Patel\\
\textit{Department of Electrical and Computer Engineering, Johns Hopkins University}\\
\textit{Baltimore, U.S.A}
}

\maketitle

\begin{abstract}

 Pseudo-label based self training approaches are a popular method for source-free unsupervised domain adaptation. However, their efficacy depends on the quality of the labels generated by the source trained model. These labels may be incorrect with high confidence, rendering thresholding methods ineffective. In order to avoid reinforcing errors caused by label noise, we propose an uncertainty-aware mean teacher framework which implicitly filters incorrect pseudo-labels during training. Leveraging model uncertainty allows the mean teacher network to perform implicit filtering by down-weighing losses corresponding uncertain pseudo-labels. Effectively, we perform automatic soft-sampling of pseudo-labeled data while aligning predictions from the student and teacher networks. We demonstrate our method on several domain adaptation scenarios, from cross-dataset to cross-weather conditions, and achieve state-of-the-art performance in these cases, on the KITTI lidar target dataset.

\end{abstract}

\section{Introduction}

Perception is an important part of autonomous driving systems, with navigation and decision making relying heavily on the ability of the vehicle to correctly localize and classify the objects around it. Recent pure lidar-based 3D object detectors have proven to perform extremely well on large public datasets and have topped their challenge leaderboards \cite{yan2018second,shi2019pointrcnn,lang2019pointpillars}. 

The recent years have seen the publication of several large autonomous driving lidar datasets  \cite{KITTI,nuscenes2019,waymo,cadc}. Although containing similar scenes of roads, pedestrians, and vehicles, they tend to differ from each other in terms of pointcloud density, the average size of lanes, as well as the types of vehicles present \cite{wang2020train}. This is due to the fact that these datasets are collected using different types of lidar sensors in varying locations around the world, and at times, under varying weather conditions. Particularly in the case of adverse weather such as rain, snow, or fog, lidar returns are affected by scattering, resulting in missing data in the final pointcloud \cite{lisa}.
Thus, there exists a gap in the domains of the training dataset (source) and the testing dataset (target), and 3D object detectors that generalize poorly drop in performance when evaluated on samples from the target domain.

\begin{figure}
    \centering
    \hspace{-15pt}
    \includegraphics[width=0.4\textwidth]{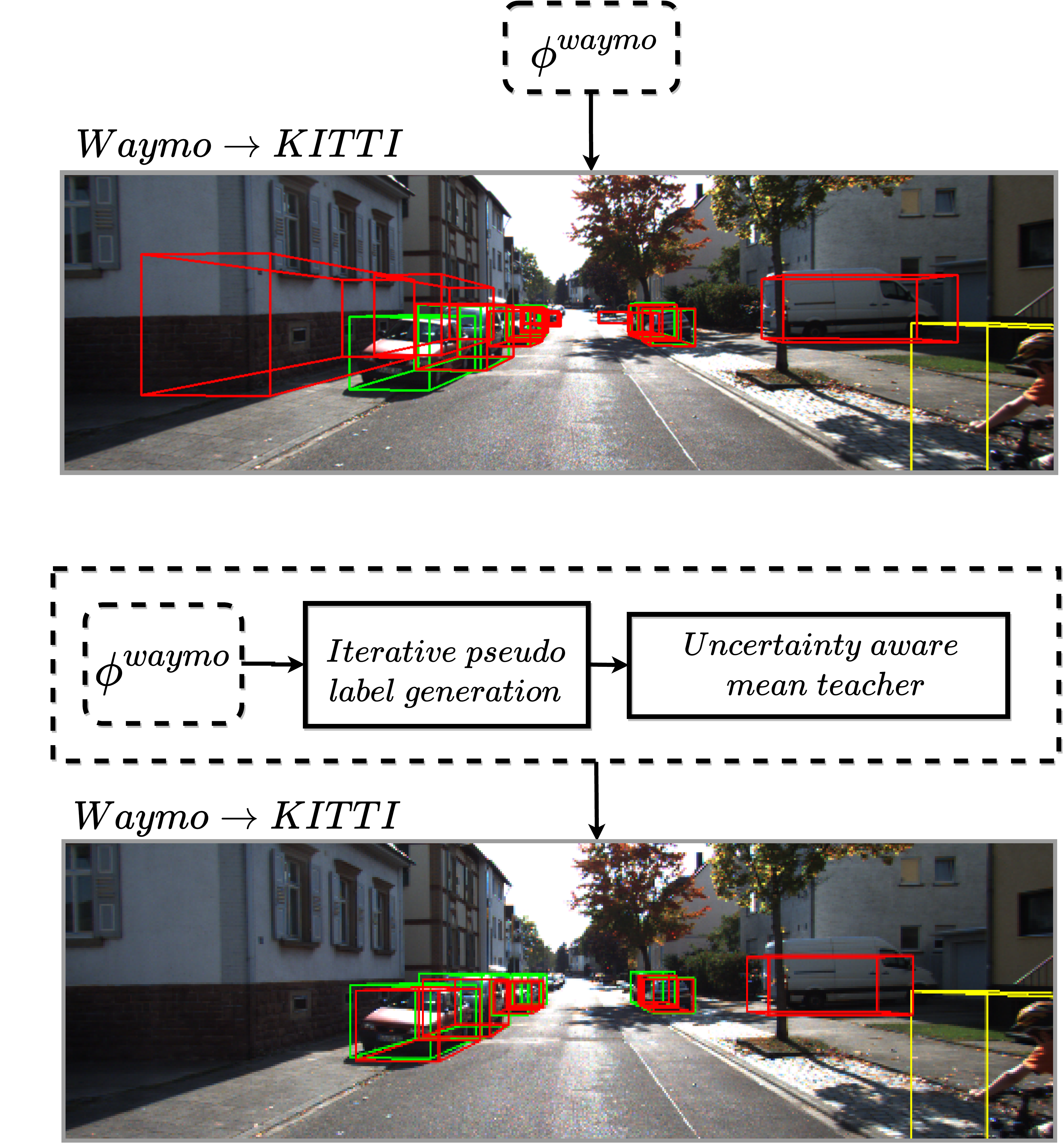}
    \caption{A comparison of the detection results of an object detection network $\phi^s$ trained on the source domain (Waymo) and tested on the target dataset (KITTI) with the results of our domain adaptation method, along with an overview of our proposed framework. The ground truth annotations are in green and the predicted results from each network are in red for the ``Car" class. }
    \label{fig:over}
\end{figure}

This poses a particular challenge in autonomous driving systems where generalization is crucial, since highway systems, driving conventions and traffic density vary  from country to country.  Naively, this problem may be mitigated by training the network with data from the various target domains. However, it would be impractical to collect and annotate every possible type of road scene from around the world. Unsupervised domain adaptation (UDA) addresses this by attempting to improve the performance of a source domain trained model on the target domain without having access to the labels of the target dataset. This has been explored on 3D data for tasks such as pointcloud classification and segmentation \cite{Qin2019PointDANAM,multimod,LIU2021211} and less extensively for 3D object detection \cite{yang2021st3d,Saltori2020SFUDA3DSU}. Recent works have also explored the semi-supervised approach by utilizing vehicle statistics from a particular area, along with the average size of objects in the target domain \cite{Wang2020TrainIG}.

Existing UDA methods require labeled source data along with the source-trained model during adaptation to the target domain \cite{yang2021st3d,scalablePseudo}. This limits applicability in scenarios where the source data is proprietary, unavailable due to privacy reasons, or too large to store. In order to address this issue we propose a source-free approach that performs domain adaptation of a network to a target domain only using a source-trained model, without the use of source domain data or labels. 

Recent domain adaptation methods for 3D object detection include semi-supervised \cite{Wang2020TrainIG} as well as source-free unsupervised approaches \cite{Saltori2020SFUDA3DSU, yang2021st3d}.  SFUDA \textsuperscript{3D} \cite{Saltori2020SFUDA3DSU}, is the first source free UDA method of this kind, but uses detection-based tracking that requires a sequence of lidar frames to estimate the quality of pseudo-labels, which limits its efficacy in real-time applications. ST3D, another unsupervised approach, obtains promising results on cross-dataset adaptation, but depends on the statistical normalization scheme from \cite{wang2020train} to surpass oracle results on the KITTI lidar dataset, which uses label data from the target domain.

We propose an unsupervised, single frame, source free domain adaptation method for 3D object detection. Using a source domain pre-trained model, we follow an iterative training scheme to generate pseudo-labels for the target domain. Although this scheme, along with the use of confidence thresholds, improves the quality of the labels, noise and incorrect labels with high confidence still exist. In order to avoid enforcing these errors while training the mean teacher network, we leverage model uncertainty to perform soft-sampling through Monte Carlo dropout-based uncertainty estimation. We explore the cross-dataset domain shift between the Waymo Open Dataset \cite{waymo} and the KITTI lidar dataset \cite{KITTI}, as well as nuScenes \cite{nuscenes2019} along with the domain shift corresponding to adverse weather conditions. An overview of our method along with a brief qualitative comparison of the result with that of a source-trained detector can be seen in Figure \ref{fig:over}.

The following are the main contributions of our work:
\begin{itemize}
    \item We propose an uncertainty-aware, self training framework for source-free unsupervised domain adaptation of 3D object detectors which implicitly selects confident samples out of a set of pseudo-labelled target data. We demonstrate our approach on the object detectors SECOND \cite{yan2018second}. 
    \item We extensively experiment on domain shifts associated with cross dataset and adverse weather scenarios, and demonstrate results on several autonomous driving lidar datasets such as the Waymo Open Dataset \cite{waymo}, the KITTI lidar dataset \cite{KITTI}, and nuScenes \cite{nuscenes2019}.

\end{itemize}

\begin{figure*}[!htp]
    \hspace{-7pt}
    \includegraphics[width=1\textwidth]{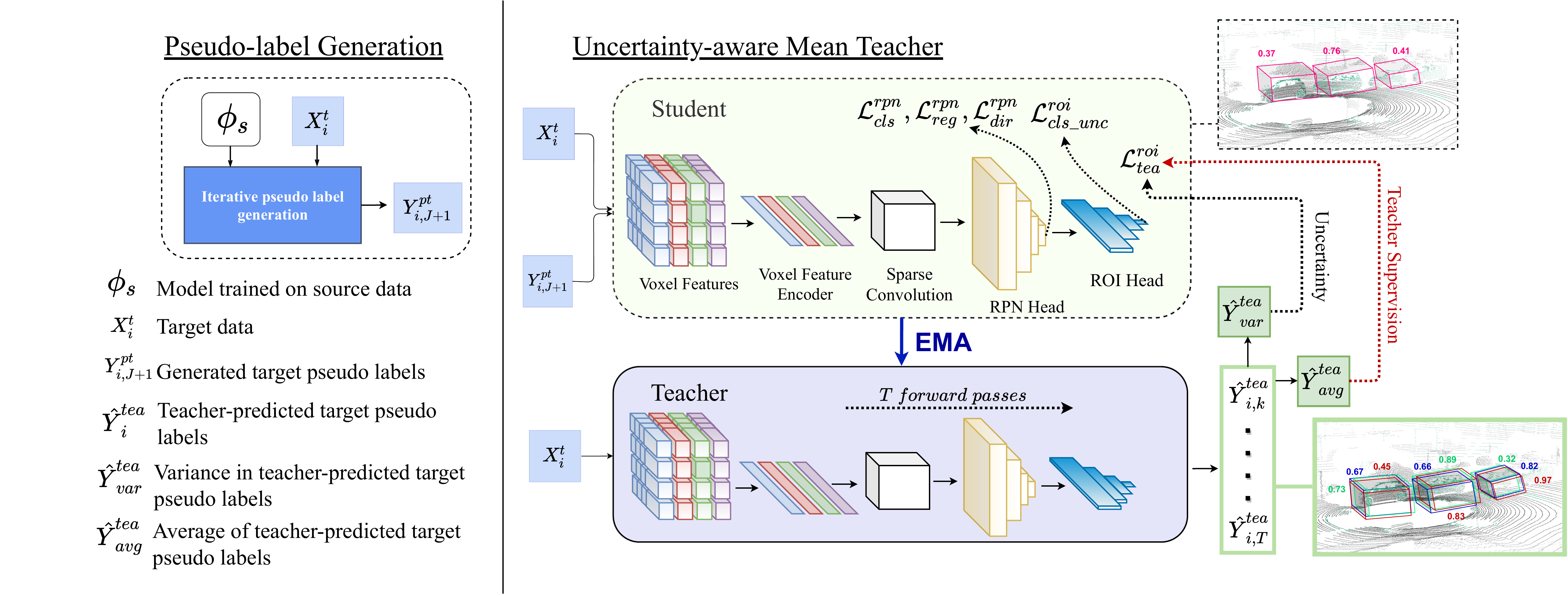}
    \caption{The two stage architecture of our proposed domain adaptation method. Initialised with a source trained model $\phi^s$, the object detector is iteratively trained with successively generated pseudo-labels to create the final set of pseudo-labels $Y_{i,J+1}^{pt}$ with which the mean-teacher network is trained. The student and teacher networks are identically initialized with $\phi^s$, and the weights of the teacher network are frozen, to be updated only via exponential moving average (EMA) of the weights from the student network. The student network is supervised by the five losses depicted (see main text for equations), including the teacher supervision loss calculated using the average and variance of the predictions from the T forward passes of the teacher network. }
    \label{fig:ps_ov}
\end{figure*}

\section{Related works}
\noindent\textbf{3D Object Detection} has been well explored in the recent years, particularly for autonomous driving scenarios, with several pure lidar-based detectors topping the challenge leaderboards of popular public datasets. These detectors aim to localize and categorize objects occurring in a 3D scene by estimating the dimensions, positions in space, and the class predictions of their bounding boxes The publication of PointNet \cite{REF:qi2017pointnet} and its successor PointNet++ \cite{REF:qi2017pointnetplusplus}, has enabled the extraction of point features by directly consuming pointclouds in an end-to-end trainable neural network. These networks form the backbone of several point-based detectors \cite{yang20203dssd,he2020structure,shi2019pointrcnn,qi2018frustum,ipod} which operate directly on the points in the Cartesian space and voxel-based detectors \cite{REF:zhou2017voxelnet,lang2019pointpillars,yan2018second}, which format the points into equally spaced 3D grids. 

PointRCNN \cite{shi2019pointrcnn} is a two stage, point-based object detector which generates and refines 3D box proposals in a bottom-up approach, through foreground-background segmentation followed by a second stage which performs bin-based box regression loss. SECOND \cite{yan2018second} is a single stage object detector which extracts voxel features from a raw pointcloud through an encoding layer, followed by sparse convolution and an RPN head which generates the detected bounding boxes. Several networks use radar and image data in addition to lidar for a multimodal approach to object detection \cite{chen2017multi,liang2019multi,qi2018frustum}.
 
\noindent\textbf{Unsupervised Domain Adaptive Object Detection} addresses the problem of the drop in performance when networks are evaluated on samples from a distribution different than that of the training set, when annotations of data in the new domain is unavailable. This has been well explored for 2D detection on images \cite{oza2021unsupervised}, including approaches such as adversarial training with domain discriminators \cite{adversarial}, image translation to bring the target domain closer to the source domain through style transfer \cite{Rodriguez2019DomainAF}, and self training approached that adapt an object detector with pseudo-labels created by source-model generated annotations \cite{Inoue2018CrossDomainWO}. Some of these ideas have been applied to object detection in the 3D domain, such as in \cite{scalablePseudo}, where Cane \etal leverage large amounts of pseudo-labeled target data along with labelled source domain data to train student networks to perform adaptation of a 3D object detector.
In \cite{yang2021st3d}, Yang \etal propose a self training domain adaptation framework that alternatively updates pseudo-labels generated by the source network and model training using curriculum data augmentation. While successful for several domain adaptation scenarios, their best results are obtained via an additional statistical normalisation step taken from \cite{wang2020train}, which uses bounding box statistics from the target labels, making it not fully unsupervised.

\noindent\textbf{Source-free domain adaptation} refers to adaptation methods which 
only use source-trained models and not the source data or labels during training. This becomes necessary when access to the source data is unavailable due to privacy, copyright, or storage restrictions. Kundu \etal \cite{Kundu2020UniversalSD} propose a two stage domain adaptation method for classification which classifies out-of-domain samples followed by adversarial alignment. There exist several approaches for 2D detection such as \cite{Li2021AFL} by Li \etal, which  utilises training with source-model generated pseudo-labels filtered by a selected by the metric of Self Entropy Descent. Recently, Saltori \etal propose SF-UDA\textsuperscript{3D},  \cite{Saltori2020SFUDA3DSU}, a source free domain adaptation framework for 3D object detection that scales pseudo-labels generated by the source-trained model to varying levels and selects the best labels through a scoring method. However, this method relies on detection-based tracking across multiple lidar frames to estimate the score for each pseudo-label.

\noindent\textbf{Mean teacher networks} are a popular method for unsupervised, semi-supervised and self-supervised approaches, particularly in object detection. Liu \etal propose \cite{liu2021unbiased}, a semi-supervised 2D object detector which jointly trains identically initialized student and teacher networks with inputs of differing levels of perturbations. The weights of the teacher network are updated by transfer from student to teacher through exponential moving average  (EMA). Cai \etal apply the mean-teacher framework for 2D cross domain detection in \cite{Cai2019ExploringOR}, and leverage inter-graph and region consistency to encourage feature space similarity between the student and teacher networks. In \cite{luo2021unsupervised}, Luo \etal propose a mean-teacher framework for unsupervised domain adaptation of 3D object detectors, and utilize point consistency, instance consistency, and neural consistency during joint learning. Although unsupervised, this approach is not source-free, and requires annotated source data during training.

\section{Proposed method}
In this section, we provide an overview of the problem and a detailed explanation of our proposed methodology. The goal is to adapt a 3D object detector trained on a source dataset to an unlabelled target dataset without the use of the source data during adaptation. 

\subsection{Preliminaries}
Consider an object detector $\phi^s$ trained on an annotated source dataset $\{(X_i^s,Y_i^s)\}_{i=1}^{N}$, where  $X^s_i$ is the $i^{th}$ sample in the set of $N$ samples, and $Y^s_i$ is the corresponding labels consisting of the location and dimensions of each bounding box. With access to this source-trained model, we adapt this detector to an unannotated target dataset $\{(X_i^t)\}_{i=1}^{M}$, where $X^t_i$ is the $i^{th}$ sample in the set of $M$ samples.

Initially, the 3D detector is trained on source data using a data augmentation method from \cite{yang2021st3d}, where objects and labels are randomly scaled with a series of factors, to give source model $\phi^s$. In the case of SECOND-iou \cite{yan2018second,yang2021st3d}, $\phi^s$ is supervised by four losses: RPN sigmoid focal classification loss, $\mathcal{L}^{rpn}_{cls}$, RPN weighted smooth $L1$ regression loss $\mathcal{L}^{rpn}_{reg}$, RPN direction classification cross entropy loss $\mathcal{L}^{rpn}_{dir}$, and ROI binary cross entropy classification loss $\mathcal{L}^{roi}_{cls}$.

We propose a framework for unsupervised domain adaptation for 3D object detection. Our approach consists of an iterative training scheme for pseudo-label generation and a student-teacher network to refine the generated pseudo-labels with Monte Carlo dropout based uncertainty estimation to mitigate label noise through soft sampling.

\begin{figure}
    \includegraphics[width=0.55\textwidth]{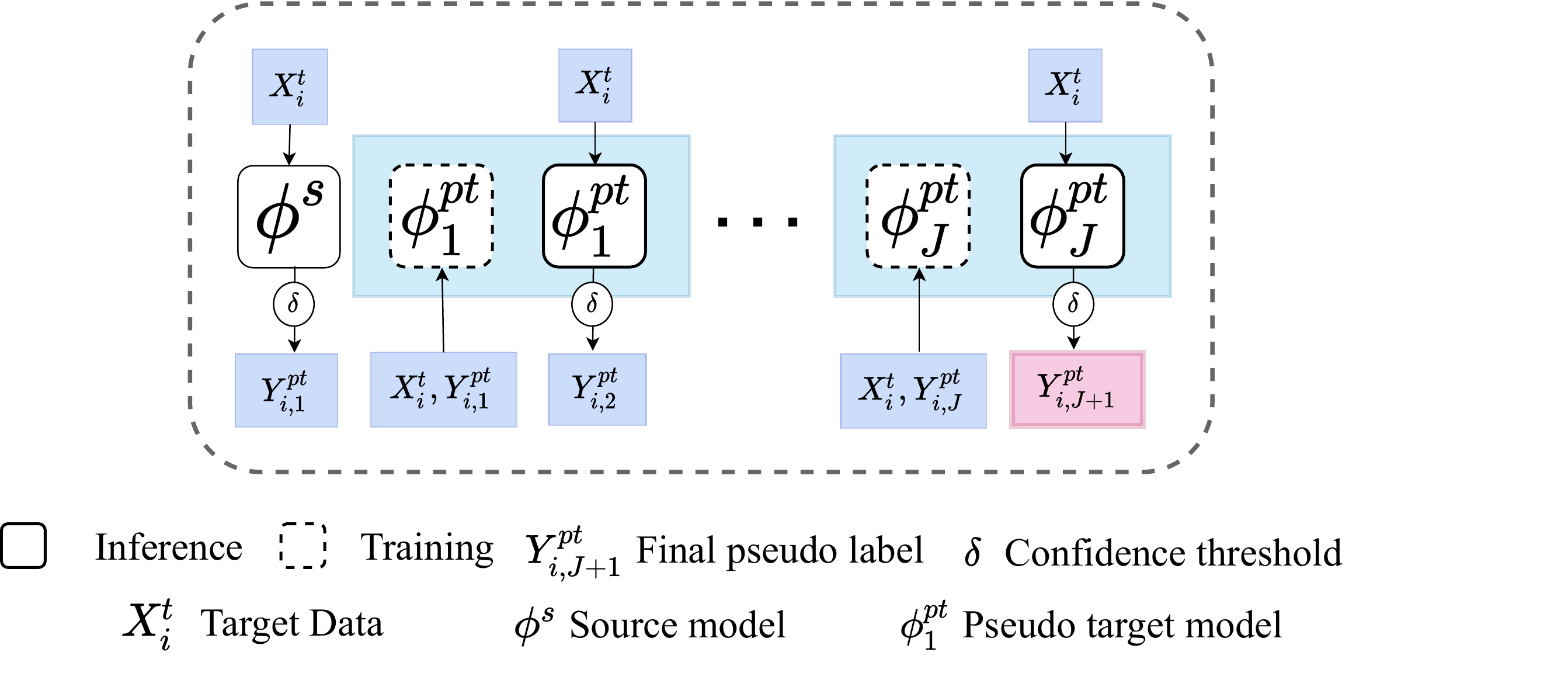}
    \caption{An expanded diagram of the iterative pseudo-label training stage. Initialized with the source trained model $\phi^s$ at each iteration, the object detector is trained on the pseudo-annotated target data, the labels of which are updated with each iteration and filtered with a threshold of $\delta$. The solid boxes indicate the inference steps where pseudo-labels are generated and the dotted-line boxes indicate training with the pseudo-labels generated in the previous iteration. }
    \label{fig:it_label}
\end{figure}

\subsection{Iterative pseudo-label generation}
\label{section:it_ps_gen}
\begin{figure}
    \hspace{-13pt}
    \includegraphics[width=0.5\textwidth]{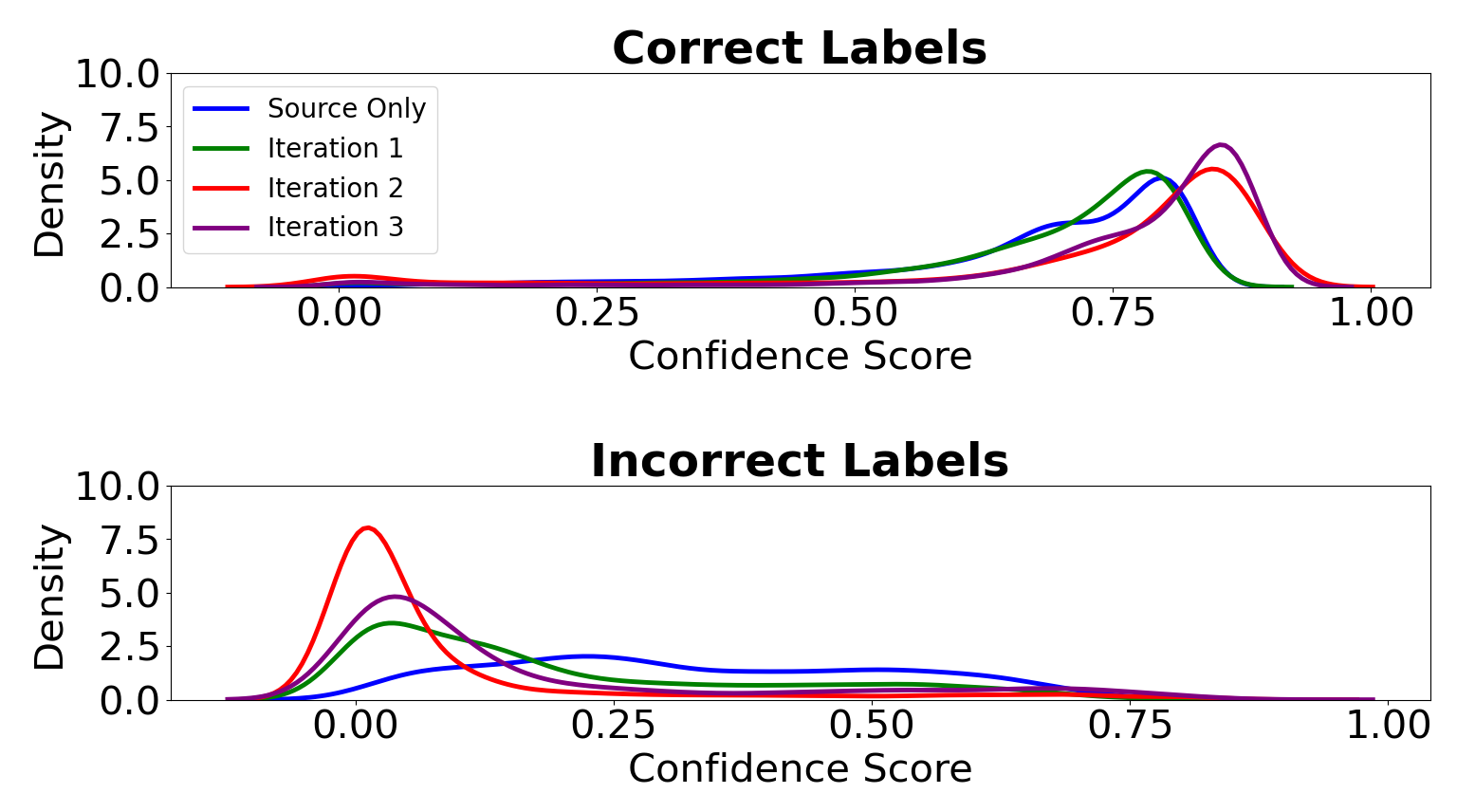}
    \caption{A density plot of correct ($IoU \geq 0.5$) and incorrect ($IoU < 0.5$) pseudo-labels according to their confidence scores for the Wayo $\rightarrow$ KITTI domain scenario. The density curve of labels generated by the source model is in blue, iteration 1 in green, iteration 2 in red, and iteration 3 in purple.  With successive iterations pseudo-label training, labels become better in quality, with higher confidence as well as higher IoU scores, as can be seen in the top graph. The peak of the last two iterations is higher and further to the left that those of the first two iterations. This indicates more correct samples of higher confidence. pseudo-labels from the source only model display higher confidence for incorrect samples, as can be seen in the bottom graph. The peaks of the red and purple curves are more defined and closer to 0, indicating incorrect pseudo-labels are of lower confidence.}
    \label{fig:iou_conf}
\end{figure}

Naively training the object detector on pseudo-labels generated by $\phi^s$ and filtered by a threshold could reinforce errors due to the fact that the source trained model may produce incorrect of higher confidence as well as correct predictions of lower confidence. We demonstrate the prevalence of label noise in Figure \ref{fig:iou_conf}, in which we plot the density of correct and incorrect pseudo-labels with respect to their confidence scored at each step for the Waymo $\rightarrow$ KITTI domain scenario. The pseudo-labels generated by the last two iterations have higher IoU with a higher confidence score as compared to those generated by the source-trained detector inferenced for the first time, and that of the first iteration. In order for thresholding to be effective, we desire correct labels to have a higher confidence score, and less confident pseudo-labels to correspond with lower IoU scores. This can be observed in the later two iterations.

In order to mitigate this label noise, we propose an iterative generation step to provide better quality pseudo-labels to the mean-teacher network, which performs further soft-sampling. 

The source-trained model is inferenced to generate pseudo-labels for target domain data $\{(X^t_i,Y^{pt}_i)\}_{i=1}^M$, where $Y^{pt}_i$ is the $i^{th}$ source-generated pseudo-label for target sample $X^t_i$. The detector $\phi$ is then initialised with the weights from $\phi^s$ and trained on the pseudo-annotated target data to give the model $\phi^{pt}$. This process is repeated $J$ number of times to give target models $\{\phi_j^{pt}\}_{j=1}^J$, each initialised with weights from $\phi^s$ and supervised with pseudo-labels $\{\{Y^{pt}_{i,j}\}_{i=1}^M\}_{j=1}^J$, filtered with a threshold $\delta$. The pseudo-labels obtained at the end of the $J^{th}$ iteration is obtained by performing inference on $\phi_J^{pt}$ and used for training the student-teacher network. An overview of this process is illustrated in Figure \ref{fig:it_label}.

\subsection{Mean teacher with Monte-Carlo dropout uncertainty}
\noindent\textbf{Mean Teacher} In order to avoid enforcing the errors present in the generated pseudo-labels during adaptation, we propose a joint learning framework based on the Mean Teacher method \cite{Tarvainen2017MeanTA} to mitigate label noise while training the object detector. This framework consists of a student network and a teacher network, both identically initialized with the source trained model weights. The student model is supervised using the generated pseudo-labels, and the weights are updated during training through backpropogation. The weights of the teacher network are gradually transferred from the student network by EMA given by 
\setlength{\belowdisplayskip}{0pt} \setlength{\belowdisplayshortskip}{0pt}
\setlength{\abovedisplayskip}{0pt} \setlength{\abovedisplayshortskip}{0pt}
\begin{equation}
    W_t \leftarrow \alpha W_t + (1 - \alpha)W_s,    
\end{equation}  where $W_t$ and $W_s$ are the weights of the teacher and student networks respectively, and $\alpha$ is the keep ratio. The weights of the teacher network are the average of the weights of the student network over multiple iterations, and thus the teacher becomes a temporal ensemble of the student. 
 
\noindent\textbf{Uncertainty Aware Student Training} The epistemic uncertainty of the teacher model is utilized by casting the network as a Bayesian Neural Network as in \cite{galBayes}, using the existing dropout layers to approximate variational inference on the network. The first moment of the predictive distribution may be calculated by performing a series of $T$ forward passes of the teacher network and averaging the results in a process called Monte-Carlo dropout \cite{galBayes}. The second moment, or predictive variance of the model may be approximated by the sample variance of the $T$ forward passes given by 
\begin{equation}
    Y^{pt}_{i,var} = \frac{\sum_{j=1}^T(Y^{pt}_{i,j} - Y^{pt}_{i,mean})^2}{T-1},
\end{equation}
where
\begin{equation}
    Y^{pt}_{i,mean} = \frac{\sum_{j=1}^T(Y^{pt}_{i,j)}}{T}.
\end{equation}

In addition to being supervised by the iteratively generated pseudo-labels through the losses present in the network, the student network is also supervised by the pseudo-labels generated by the teacher network in each epoch. The degree of model uncertainty of the teacher is represented by the variance in predictions. The teacher supervises the student with a Binary Cross Entropy loss. The pseudo-labels generated by the teacher network are obtained by computing the sigmoid of the average predictions of $T$ forward passes. The loss value for each ROI is weighted by the inverse of the computed variance. Predicted values with higher variance are thus down-weighted, effectively sampling the data to mitigate noise in the pseudo-labels. This loss can be written as  
\begin{equation}
    \mathcal{L}^{roi}_{tea} = \frac{1}{N}\sum^{i=1}_{N}\{ \mathcal{C} * \mathcal{L}_{BCE}(Y^{roi\_stu}_{pred}, Y^{roi\_tea}_{ps}) \},
\end{equation}
where $\mathcal{C}$ is the inverse of the predictive variance,  $Y^{roi}_{pred}$ is the predicted classification output of the ROI head of the student network, and $Y^{roi\_tea}_{ps}$ is the pseudo-label given by the teacher network after $T$ forward passes.

 During joint training, only the student network is supervised by the existing network losses, and $\mathcal{L}^{roi}_{cls}$ is scaled by the uncertainty weights obtained from the predictions of the teacher network. The total loss is thus given by
\begin{align*}
    \mathcal{L}_{total} &= \mathcal{L}^{rpn}_{cls} + \mathcal{L}^{rpn}_{reg}    +  \mathcal{L}^{rpn}_{dir} + \mathcal{L}^{roi}_{cls\_unc} + \mathcal{L}^{roi}_{tea},
\end{align*}
where
\begin{equation}
    \mathcal{L}^{roi}_{cls\_unc} = \frac{1}{N}\sum^{i=1}_{N}(\mathcal{C}*\mathcal{L}_{BCE}(Y^{roi}_{pred}, Y^{roi}_{ps})),
\end{equation}
and $N$ is the total number of valid ROIs, $Y^{roi}_{pred}$ is the predicted classification output of the ROI head of the student network, and $Y^{roi}_{ps}$ is the pseudo ground truth label of the ROIs generated by the teacher network. 

\begin{table*}[!htp]
\centering
\caption{A comparison of results for 3 domain shift scenarios from the proposed method with that of a recent domain adaptive 3D object detector ST3D \cite{yang2021st3d}, a semi-supervised domain adaptive technique statistical normalization (SN) \cite{wang2020train}, the source-model performance, and the oracle performance of the object detector, i.e. trained in a fully supervised manner with target data. Our proposed framework outperforms the compared method in most categories across all the explored domain shifts. }
\vspace{10pt}
\resizebox{0.7\textwidth}{!}{%
\begin{tabular}{c|c|ccc} 
\hline 
\toprule[1pt]\midrule[0.3pt]    
\multirow{2}{*}{Domain shift}                   & \multirow{2}{*}{Method} & \multicolumn{3}{c}{mAP (BEV/3D)}                                           \\ \cline{3-5} 
                                                &                         & \multicolumn{1}{c}{Easy} & \multicolumn{1}{c}{Moderate} & Hard              \\ \hline
\multirow{5}  {*}{Waymo $\rightarrow$ KITTI}      & Source Only             & 82.57 / 46.66         & 66.70 / 39.86           & 59.02 / 35.03 \\
                                                & SN                      &   86.86 / 73.23         &  72.46 / 56.36              &  68.63 /  54.25      \\
                                                & ST3D                    & 87.97 / 67.47         & 77.36 /  59.17          & 76.09 / 56.73 \\
                                                & Proposed                & \textbf{88.60} / \textbf{75.35}         & \textbf{77.79} / \textbf{64.56}          &  \textbf{76.48} / \textbf{61.02} \\ \cline{2-5} 
                                                & Oracle                  & 89.55 / 84.86         & 79.29 / 68.93           & 78.26 / 67.38 \\ \hline

\toprule[0.1pt]\midrule[0.1pt]
\multirow{5}  {*}{nuScenes $\rightarrow$ KITTI}   & Source Only             & 53.15 / 18.37         & 46.21 / 17.31           & 43.64 / 16.09 \\
                                                & SN \cite{wang2020train}                &   39.43 / 22.03   &           31.88 / 18.51    &      31.61 / 18.04     \\
                                                & ST3D \cite{yang2021st3d}                    & \textbf{86.39} / 58.24             & \textbf{74.61} /  43.13              & 70.95 / 39.46 \\
                                                & Proposed      &   82.50 / \textbf{62.09}   &         72.33 / \textbf{47.91}        &      \textbf{70.99} /  \textbf{42.36}  \\ \cline{2-5} 
                                                & Oracle                  & 89.55 / 84.86         & 79.29 / 68.93           & 78.26 / 67.38 \\ \hline

\toprule[0.1pt]\midrule[0.1pt]
\multirow{5}  {*}{Waymo $\rightarrow$ KITTI-rain} & Source Only             & 60.36 / 31.61         & 40.66 / 23.73           & 38.74 / 21.12 \\
                                                & SN                      &  68.23 / 40.67            &  47.99 / 30.95    &       46.32 / 28.05                        \\
                                                & ST3D                    & \textbf{76.45} / 52.16         & \textbf{57.21} / 37.57           & \textbf{53.24} / 35.20 \\
                                                & Proposed                &     74.08 / \textbf{53.27}        &    55.89 / \textbf{38.99}           &    53.17 / \textbf{35.25}  \\ \cline{2-5} 
                                                & Oracle                  & 78.14 / 63.89         & 59.79 / 47.22           & 51.45 / 39.95 \\ 
\midrule[0.1pt]\toprule[1pt]
\end{tabular}
\label{comp}
}
\end{table*}

\begin{figure}
    \hspace{-5mm}
    \includegraphics[width=0.5\textwidth]{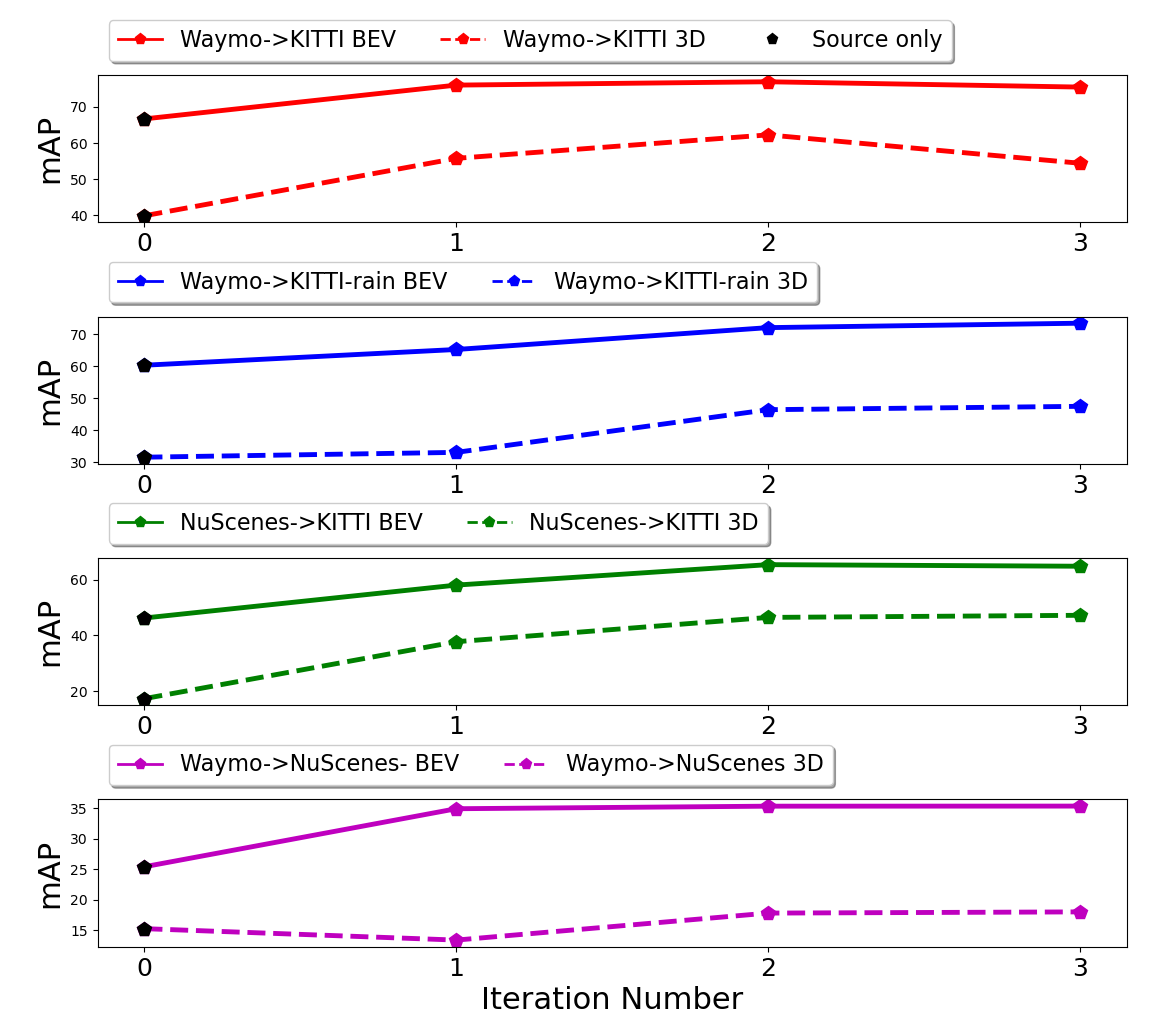}
    \caption{A plot of mean average precision (mAp) for the ``Moderate" category and ``Car" class of the object detectors $\{\phi_j^{pt}\}_{j=1}^J$ at each iteration $j$ of the pseudo-label generation process for each domain shift scenario. With each successive iteration until $J\approx3$, there is an improvement in the performance of the detectors, with performance eventually saturating.}
    \label{fig:pseudo_map}
\end{figure}

\begin{table}[!htp]
\centering
\caption{A comparison of results for Waymo $\rightarrow$ nuScenes domain shift scenario from the proposed method with that of a recent domain adaptive 3D object detector ST3D \cite{yang2021st3d}, statistical normalization (SN) \cite{wang2020train}, the source-model performance, and the oracle performance of the object detector. Our proposed framework outperforms the compared method in most categories. }
\vspace{10pt}
\resizebox{0.45\textwidth}{!}{%
\begin{tabular}{c|c|ccc}

\toprule[1pt]\midrule[0.3pt]
{Domain shift}                   & {Method} & \multicolumn{3}{c}{mAP (BEV/3D)}                                           \\ \cline{3-5}

\hline
\multirow{5}  {*}{Waymo $\rightarrow$ nuScenes}   & Source Only             & 25.39 / 15.25                                     \\
                                                & SN \cite{wang2020train}                     & 33.23 / 18.57                                                      \\
                                                & ST3D \cite{yang2021st3d}                   &   29.07 / 15.52                                  \\
                                                & Proposed                &    \textbf{35.10} / \textbf{21.05}                                                  \\ \cline{2-5} 
                                                & Oracle                  & 49.32 / 32.65                                      \\ 
\midrule[0.1pt]\toprule[1pt]

\end{tabular}
}
\label{comp}

\end{table}

\begin{table*}[]
\centering
\caption{A comparison of mean average precision (mAP) values for the 3D Moderate category of a mean teacher framework with and without uncertainty aware weighing of regions during teacher supervision. As can be observed, performance in most categories is increased by several points when the network is trained to be uncertainty-aware.}
\vspace{5pt}
\resizebox{0.65\textwidth}{!}{%
\begin{tabular}{c|ccc|ccc}

\midrule[1pt]\toprule[0.1pt]
\multirow{2}{*}{Domain shift} & \multicolumn{3}{c|}{Mean teacher}                         & \multicolumn{3}{c}{Uncertainty-aware mean teacher}        \\ \cline{2-7} 

                              & Easy              & Moderate            & Hard              & Easy              & Moderate            & Hard              \\ \hline
nuScenes $\rightarrow$ KITTI                &  52.63 &  42.04 &  40.70 & 62.09   &        47.91        &      42.36   \\
Waymo $\rightarrow$ KITTI                   &  75.28 &  58.11 &  56.55 &  75.35 &  64.56 &  61.02 \\
Waymo $\rightarrow$ KITTI-rain              &  52.03 &  35.90 &  30.73 &    53.27 & 38.99 & 35.25        \\
Waymo $\rightarrow$ nuScenes                &         -          &    17.08         &        -           &              -     &     21.05     &          -         \\ \midrule[0.1pt]\toprule[1pt]
\end{tabular}
}
\label{ref:mt_unc}
\end{table*}

\begin{figure}
    \centering
    \hspace{-5mm}
    \includegraphics[width=0.45\textwidth]{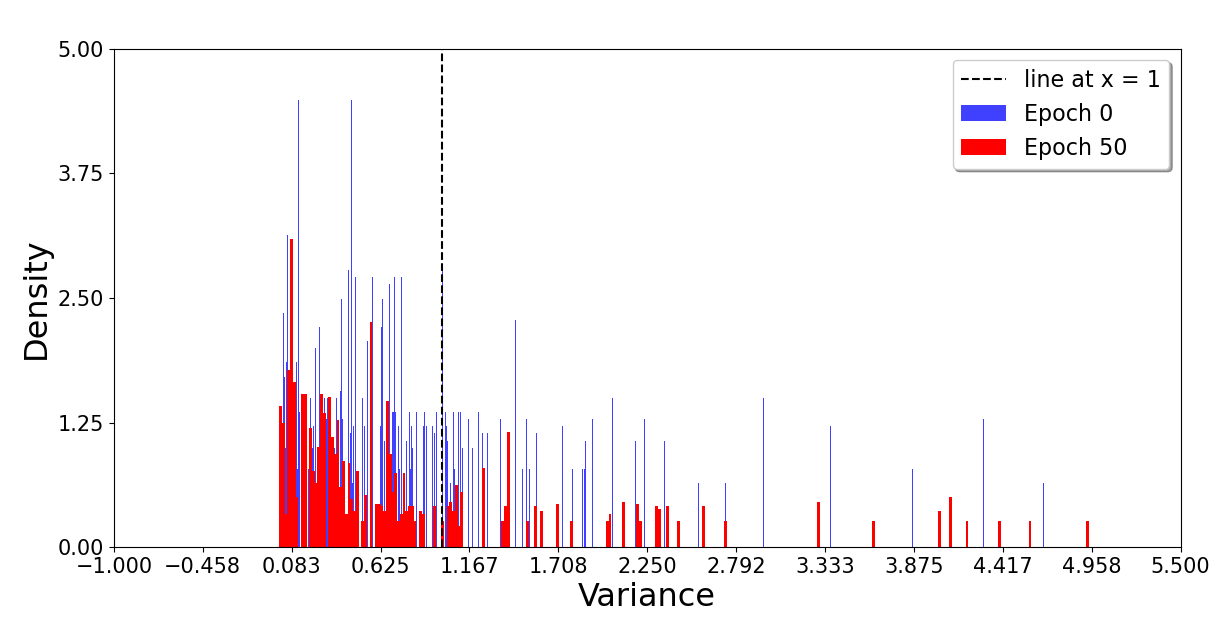}
    \caption{A density plot of model variance of incorrectly labeled ROIs at the initial and final epochs. At the beginning of training, the model initialized with source-model weights has a larger number of incorrect labels with low uncertainty ($variance <1$), shown in blue. At the final epoch, the distribution of the incorrect samples (shown in red) moves to the right, with a fewer number of labels with low values of variance.  }
    \label{fig:unc_hist}
\end{figure}

\begin{table}[!h]
\centering
\caption{The left column shows a comparison of mean average precision (mAP) values of the object detection network for the ``Moderate" difficulty at each iteration of the pseudo-label generation process for the nuScenes $\rightarrow$ KITTI domain scenario. As is apparent in the table, with each successive iteration, the quality of the pseudo-labels increases, as does the performance of the self-training method. The left column demonstrates a similar effect, with the performance of the uncertainty-aware mean teacher framework with each iteration of pseudo labels.}
\vspace{5pt}
\resizebox{0.47\textwidth}{!}{%
\begin{tabular}{c|c|c}
\midrule[1pt]\toprule[0.1pt]
Iteration & Self training &    Uncertainty-aware mean teacher       \\ \cline{1-2} 
\hline
source only                   & 46.21 / 17.31 & 60.03 / 40.48\\
iteration 1                   & 58.59 /  37.71 &  62.89 / 44.01 \\
iteration 2                   & 65.33 / 46.42 & 74.19 / 42.77\\
iteration 3                   & 64.76 / 47.17 &  72.33 / 47.91\\ 
\midrule[0.1pt]\toprule[1pt]
\end{tabular}

\label{ref:it_ps}
}
\end{table}

\section{Experiments}
In this section, we explain the details of the experiments carried out and conduct an analysis of model uncertainty.

\subsection{Experimental setup}

\noindent\textbf{Datasets}
We demonstrate our method on three well-known lidar detection datasets -- Waymo\cite{waymo}, KITTI \cite{KITTI}, and nuScenes\cite{nuscenes2019}. The Waymo Open Dataset is collected from six cities across the United States, and consists of X training samples and Y validation samples of which we use a subset of X' and Y' samples for training and testing, respectively. The KITTI dataset is one of the earliest publicly available lidar datasets for scene understanding, and is much smaller in size with 3712 training samples and 3769 validation samples, which we use for testing. The nuScenes dataset consists of 34,149 samples in a 28130/6019 training/validation split. We address the domain shifts of cross-dataset detection when going from label-rich to label-poor scenarios, by exploring the following scenarios:  Waymo $\rightarrow$ KITTI, nuScenes $\rightarrow$ KITTI, and Waymo $\rightarrow$ nuScenes. 

We also address the domain shift caused by adverse weather conditions, by simulating rain on the KITTI dataset using the physics-based lidar weather simulation algorithm proposed in \cite{lisa}. Sampling from a range of rain rates from $0 mm/hr$ to $100 mm/hr$ to reflect realistic adverse weather conditions, each sample is augmented with the artifacts that occur when lidar data is captured in rainy weather.  We denote this domain shift as Waymo $\rightarrow$ KITTI-rain.
 
\noindent\textbf{3D object detection network} We demonstrate the proposed method on the 3D object detector SECOND-iou \cite{yan2018second,yang2021st3d}. SECOND is a voxel-based, single stage 3D object detection network consisting of a pointcloud-grouping step to create voxels and their corresponding coordinates, a feature extractor based on \cite{zhou2018voxelnet}, which applies a fully connected layer, batch normalization , and ReLU on each voxel, a sparse convolution layer, followed by a region proposal network (RPN) provides the category and dimensions of the bounding boxes predicted. SECOND-iou is a slightly modified version of this network proposed by Yang \etal in \cite{yang2021st3d}, which has an additional refinement ROI-head.  This is a single stage detector, and the entire network is used for the mean-teacher framework.

\subsection{Implementation details} We implement the proposed framework on SECOND-iou using the codebase OpenPCDet \cite{openpcdet2020}, a popular open-source repository for 3D object detection networks. We also refer to code from \cite{yang2021st3d} for their implementation of the extra ROI head. During iterative pseudo-label generation step as well as the mean-teacher training step, we use a series of confidence threshold of $\delta \in \{0.1,0.6,0.8\}$. Due to computational limitations, we run our method with a batch size of 16, although higher batch sizes are used by \cite{yang2021st3d}. During source model pre-training, we utilize the random object scaling data augmentation procedure proposed by \cite{yang2021st3d} along with standard augmentation steps such as global scaling and random object rotation and train for 50 epochs. The teacher network is trained with a series of $T=15$ forward passes, and a keep ratio  of $\alpha=0.999$ for EMA. During mean teacher training, the entire network is trained for 50 epochs.

\section{Results}
In this section, we present the results of our proposed framework, and compare \footnote{To ensure a fair comparison across all evaluation categories, we re-implement the comparative methods with the same batch size, number of epochs, and other hyperparameters as our models.} it with recent methods for domain adaptation for 3D object detectors, namely Statistical Normalization (SN) \cite{wang2020train} and ST3D \cite{yang2021st3d}, along with the source-only and oracle performances of the object detector.
\subsection{Evaluation metrics}
We evaluate the model on the official metrics of the  KITTI dataset \cite{KITTI}, which divides each object in each sample into 3 categories, Easy, Moderate, and Hard based on the distance of the object from the sensor and amount of occlusion. Mean average precision is calculated for the bird's eye view (BEV) as well as for the entire 3D bounding box, with an IoU threshold of $0.7$. For the Waymo and nuScenes datasets, we consider a full $360^{\circ}$ view whereas only the front view is considered for samples from the KITTI dataset. Evaluation of the networks is performed on the ``Car" class in the KITTI dataset, with the equivalent class in Waymo and nuScenes being ``Vehicle" and ``car", respectively.

                             


\begin{figure*}
    \centering
    \includegraphics[width=0.9\textwidth]{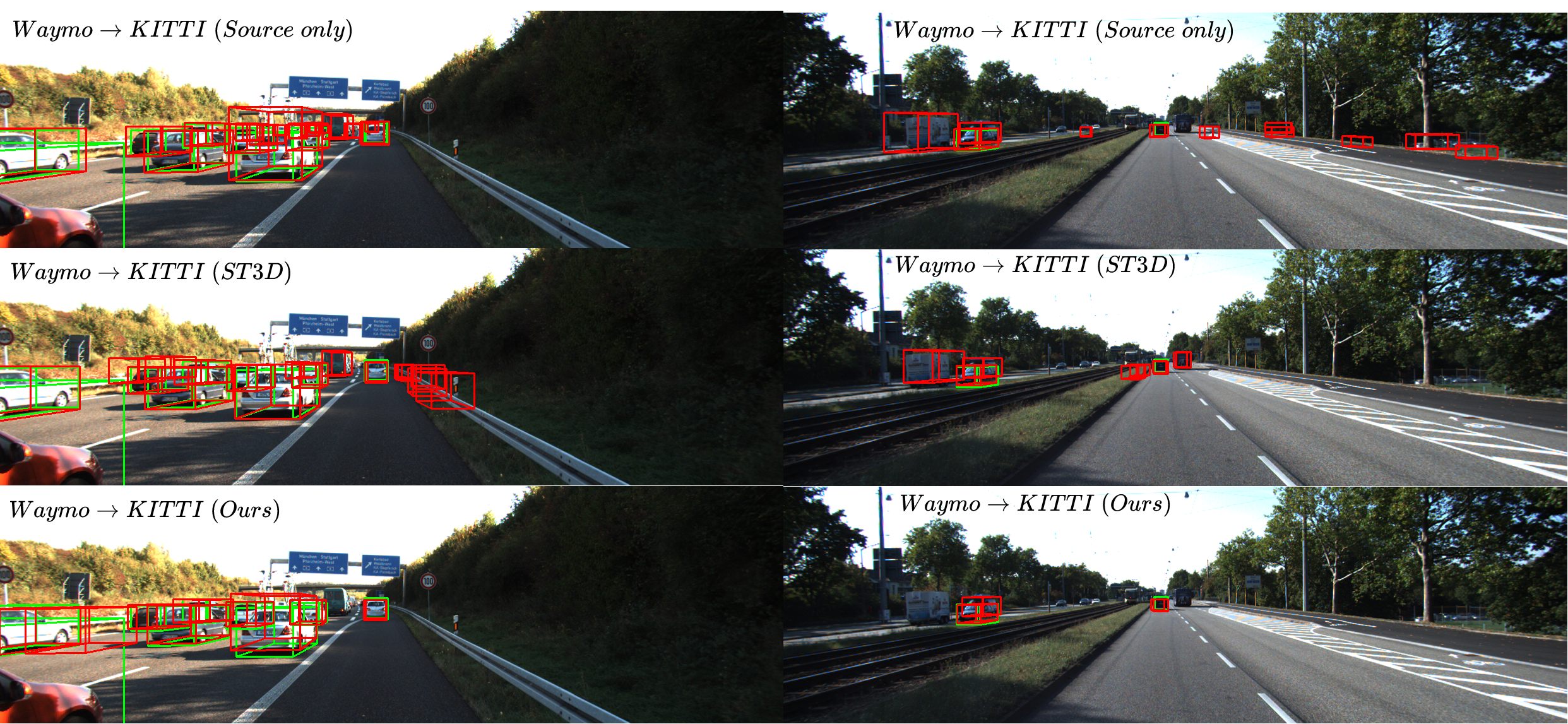}
    \caption{A qualitative comparison our results against that of the source trained model and that of ST3D \cite{yang2021st3d} for the Waymo $\rightarrow$ KITTI. The first row shows samples with bounding boxes predicted by the source trained model, followed by the results of ST3D and the proposed method in the second and third rows. The ground truth bounding boxes are in green and the predictions are in red. (Best viewed zoomed in and in color.)  }
    \label{fig:viz1}
\end{figure*}

\subsection{Comparison with SOTA methods}

\noindent\textbf{Quantitative results} We compare the quantitative results of our method against that of ST3D \cite{yang2021st3d}, which uses the same base object detection network (SECOND-IoU) as well as with the semi-supervised method of statistical normalization proposed by Wang \etal in \cite{wang2020train}, well as the oracle performance of the network, obtained by fully supervising the detector with the target dataset during training. These results for each domain shift scenario may be seen in Table \ref{comp}. We obtain the highest mean average precision in almost all categories across all domain shifts. Particularly in the Waymo $\rightarrow$ KITTI and Waymo $\rightarrow$ nuScenes scenarios, our method outperforms the rest. 

\noindent\textbf{Qualitative results} We also compare the visual quality of the results against that of the source-only network and ST3D \cite{yang2021st3d} for the Waymo $\rightarrow$ KITTI domain scenario. As can be seen in Figure \ref{fig:viz1}, both the source trained model and ST3 suffer from incorrect samples with high confidence, negatively affecting the precision. Our method avoids this, and also demonstrated better localization.


\subsection{Analysis of model uncertainty}
We examine the model uncertainty of the object detector during adaptation at the beginning and end of training. We evaluate the pseudo labels by comparing the pseudo class annotation with the ground truth annotation for each detected object. We then plot the variance of the teacher network for the associated region of interest (ROI) of incorrect pseudo labels.   We plot the distributions of these variances in Figure \ref{fig:unc_hist}. As is apparent in the figure, by the end of training, the model learns to better identify incorrect pseudo labels, since fewer incorrect samples have low variance, ensuring that they are down-weighed during training

\subsection{Ablation study}
We conduct several ablation experiments to examine the contribution of each part of the architecture to the performance of the object detector. 

\noindent\textbf{Iterative pseudo-label generation} As mentioned in Section \ref{section:it_ps_gen}, we use an iterative training strategy to initially generate pseudo-labels to train the mean-teacher network. With confidence thresholds $0.1<\delta<0.8$ at each iteration, this process helps to filter low confidence pseudo-labels at several levels. In the second column of Table \ref{ref:it_ps}, we compare the performance of the detector when trained with source-only generated pseudo-labels and at each subsequent iteration of the pseudo-label generation process for the nuScenes $\rightarrow$ KITTI domain setting. As observed in the table, the performance of simple self-training in mAP improves with each iteration of pseudo-label generation. This can likewise be observed in the other domain shifts in Figure \ref{fig:pseudo_map}, here the mean average precision (mAP) values of each self-trained object detector $\{\phi_j^{pt}\}_{j=1}^J$ in the ``Moderate" difficulty category for each iteration are plotted for each domain shift scenario. As observed in the figure, the precision values increase with each iteration before either saturating or decreasing. We observe that most models reach this point at $J \approx 3 $. We also compare the performance of the uncertainty aware mean teacher framework in this setting when trained with each set of pseudo-labels in the third column of Table \ref{ref:it_ps}. It is apparent that the mean teacher framework benefits from the iterative pseudo-label generation, due to the improved quality. 

As mentioned in Section 3.2, Figure \ref{fig:iou_conf} shows that pseudo-labels generated in later iterations tend to be more confident as well as more correct (i.e. higher IoU with ground truth labels) when compared to pseudo-labels generated by the source trained model and the first iteration model.

\noindent\textbf{Uncertainty-aware supervision}
In order to examine the role of uncertainty-aware teacher supervision of the student network, we compare the Moderate 3D performance of the mean-teacher framework with and without down-weighing losses obtained from the variance of the $T$ forward passes of the network. From Table \ref{ref:mt_unc}, one can observe that there is significant performance improvement in most categories of the explored domain shifts. The mean teacher framework clearly benefits from being made uncertainty-aware through Monte-Carlo Dropout, and significantly mitigates noise not only on source-model generated pseudo labels, but the iteratively refined labels as well.

\section{Conclusion}
We proposed an uncertainty-aware mean teacher framework for domain adaptive 3D object detection, which improves upon naive pseudo-label based self training methods through the  mitigation of label noise by down weighing samples the teacher model is uncertain about. We show improved performance across four domain shift scenarios, outperforming the improvement demonstrated by recent unsupervised and semi-supervised domain adaptation methods.  

{\small
\bibliographystyle{ieee_fullname}
\bibliography{egbib}
}

\clearpage



\section*{A. Algorithms}

In this section, we detail the training procedures and algorithms for our two stage framework for source free unsupervised domain adaptive 3D object detection. Algorithm \ref{alg:1} details the iterative pseudo label generation procedure illustrated in Figure 3 of the main paper. Algorithm \ref{alg:cap} explains the steps involved in training the uncertainty aware mean teacher with the pseudo-labels generated in the previous stage.

\begin{algorithm}[H]
\caption{Iterative pseudo-label generation}\label{alg:1}
 \algorithmicrequire{  source trained model $\phi^s$, unannotated target data} \\
 \algorithmicensure{  $\{Y^{pt}_{i,J}\}_{i=1}^M$ }
 \begin{algorithmic}
    \STATE - Perform inference on $\phi^s$ to obtain $\{Y^{pt}_{i,0}\}_{i=1}^M$
    \STATE - Threshold with $\delta[0]$
      \FOR{$1 \leq j \leq J$}
     \STATE{- Init $\phi$ with $\phi^s$}
     \STATE - train $\phi$ with target data of $M$ samples annotated with $\{Y^{pt}_{i,j-1}\}_{i=1}^M$
     \STATE - Obtain trained model $\{\phi_j^{pt}\}$
     \STATE - Perform inference on $\{\phi_j^{pt}\}$ to obtain pseudo labels $\{Y^{pt}_{i,j}\}_{i=1}^M$
     \STATE - Threshold with $\delta[j]$
     
     \ENDFOR
     
     \STATE - Output pseudo labels = $\{Y^{pt}_{i,J}\}_{i=1}^M$
 \end{algorithmic}
\end{algorithm}
\vfill
\section*{B. Additional Qualitative Results}
\begin{figure*}
    \centering
    \includegraphics[width=\linewidth]{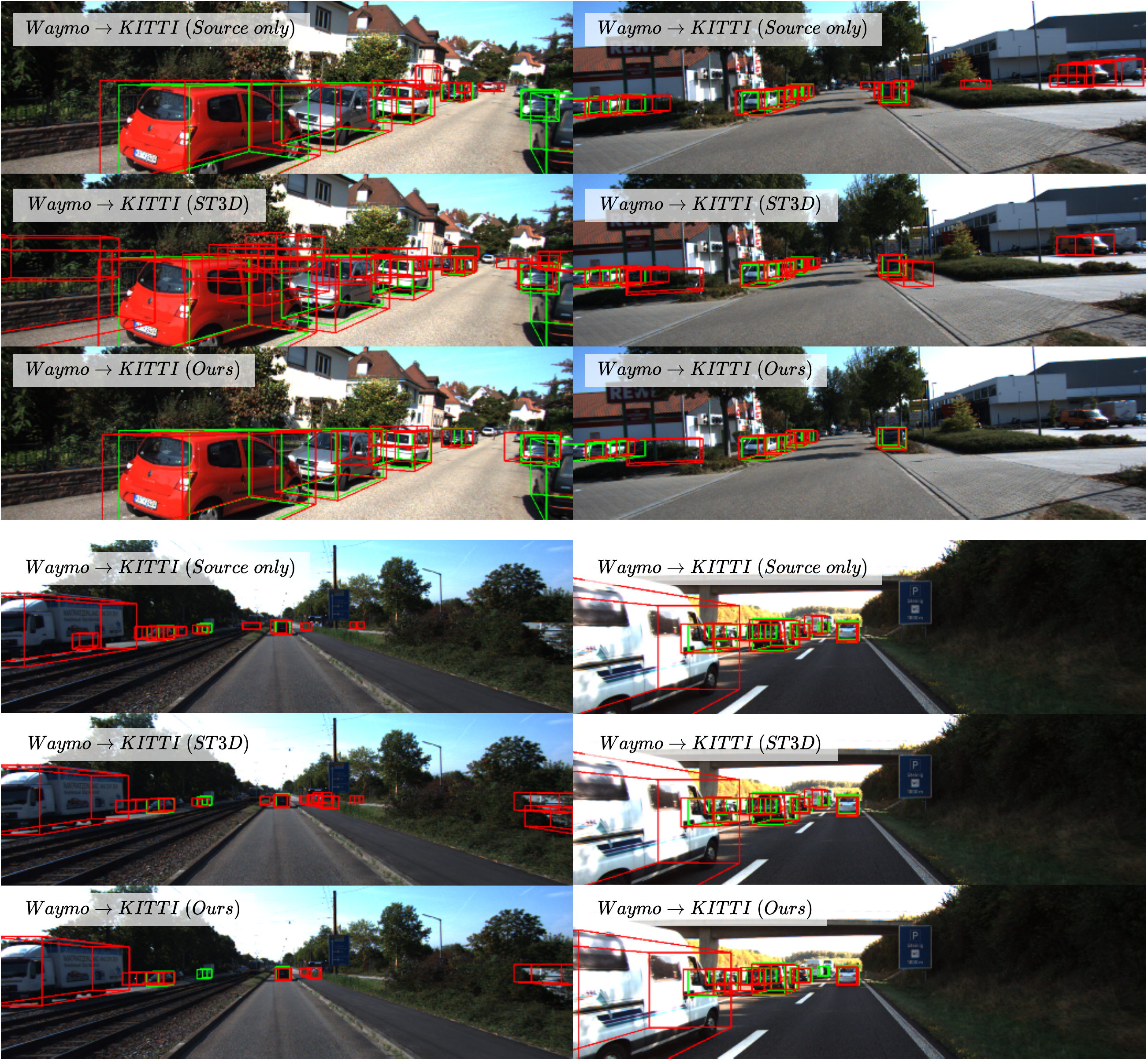}
    \caption{A qualitative comparison of detection results from the source-trained model, ST3D \cite{yang2021st3d}, and our proposed framework. The ground truth bounding boxes for the ``Car'' class are in green and the detected boudning boxes are in red. The first and fourth rows show the results from the source trained model, the second and fifth from \cite{yang2021st3d}, while the third and final rows show the results of the proposed method. }
    \label{fig:qual}
\end{figure*}

In this section, we provide further qualitative comparisons with recent domain adaptation method \cite{yang2021st3d} as well as the source trained object detector. In Figure \ref{fig:qual}, we demonstrate the improved localization of the proposed method as compared to the source-trained network and that of ST3D \cite{yang2021st3d}. Both these methods suffer from incorrect predictions of high confidence, resulting in spurious detections and lower precision.

\begin{algorithm}[H]
\caption{Training the uncertainty-aware mean teacher network}\label{alg:cap}
 \algorithmicrequire{  source trained model $\phi^s$, unannotated target data,pseudo labels $\{Y^{pt}_{i,J}\}_{i=1}^M$} \\
 \algorithmicensure{Predictions}
 \begin{algorithmic}
    \STATE - Initialize teacher model with $\phi^s$
    \STATE - Freeze $autograd$ for teacher network.
    \STATE - Initialize student model with $\phi^s$

      \FOR{$1 \leq epoch \leq num\_epochs$}
      \STATE - Train student net with target data annotated with pseudo-labels $\{Y^{pt}_{i,J}\}_{i=1}^M$
      \STATE - init $pred\_teacher$
      \FOR{$1 \leq i \leq T$}
      
        \STATE - Perform a forward pass on the teacher network
        \STATE - Append output to $pred\_teacher$

      \ENDFOR
      
      \STATE - Transfer weights from student network to teacher network through exponential moving average (EMA).
      
      \STATE - Compute losses $\mathcal{L}^{rpn}_{cls}$, $\mathcal{L}^{rpn}_{reg}$,  $\mathcal{L}^{rpn}_{dir}$, and $\mathcal{L}^{roi}_{cls}$ from the output of the student network.
      
      \STATE - Compute $\hat{Y}_{var}^{tea}$ and $\hat{Y}_{avg}^{tea}$ from $pred\_teacher$. 
      
      \STATE - Compute loss weight $C=clip(\frac{1}{\hat{Y}_{var}^{tea}},10^{-5},1)$
      
      \STATE - Compute $\mathcal{L}^{roi}_{cls} \leftarrow C\times \mathcal{L}^{roi}_{cls}$ 
      
      \STATE - Compute $\mathcal{L}^{roi}_{tea}$
      
      \STATE - Backpropagate losses through student network.

     \ENDFOR
     
     \STATE - Output bounding box predictions.
 \end{algorithmic}
\end{algorithm}


\end{document}